\documentclass{article}
\usepackage{spconf,amsmath,amssymb,graphicx}
\usepackage{mathtools}
\usepackage{float}
\usepackage{bm}
\usepackage{multirow}
\usepackage{color}
\usepackage{tikz}

\newcommand{\kl}[1]{\textcolor{red}{#1 (KL)}\ }
\newcommand{\bs}[1]{\textcolor{blue}{#1 (BS)}\ }

\renewcommand{\kl}[1]{}
\renewcommand{\bs}[1]{}

\title{Multitask training with unlabeled data \\ for end-to-end sign language fingerspelling recognition}
\name{Bowen Shi, Karen Livescu}
\address{Toyota Technological Institute at Chicago\\
\{bshi, klivescu\}@ttic.edu}
\begin{document}
%
\maketitle
\begin{abstract}
We address the problem of automatic American Sign Language fingerspelling recognition from video.  
Prior work has largely relied on frame-level labels, hand-crafted features, or other constraints, and has been hampered by the scarcity of data for this task.
We introduce a model for fingerspelling recognition that addresses these issues.  The model consists of an auto-encoder-based feature extractor and an attention-based neural encoder-decoder, which are trained jointly.
The model receives a sequence of image frames and outputs the fingerspelled word, without relying on any frame-level training labels or hand-crafted features. In addition, the auto-encoder subcomponent makes it possible to leverage unlabeled data to improve the feature learning. The model achieves 11.6\% and 4.4\% absolute letter accuracy improvement respectively in signer-independent and signer-adapted fingerspelling recognition over previous approaches that required frame-level training labels. 
\end{abstract}
\section{Introduction}
\label{sec:intro}

Automatic recognition of sign language from video could enable a variety of services, such as search and retrieval for Deaf social and news media (e.g., \texttt{deafvideo.tv, aslized.org}). Sign language recognition involves a number of challenges.  
For example, sign languages each have their own grammatical structure with no built-in written form; ``transcription" of sign language with a written language is therefore a translation task.  In addition, sign languages often involve the simultaneous use of handshape, arm movement, and facial expressions, whose related computer vision problems of articulated pose estimation and hand tracking still remain largely unsolved. 
Rather than treating the problem as a computer vision task, many researchers 
have therefore chosen to address it as a linguistic task, with speech recognition-like approaches.

In this paper, we focus on recognition of fingerspelling, a part of ASL in which words are spelled out letter by letter (using the English alphabet) and each letter is represented by a distinct handshape. Fingerspelling accounts for 12 - 35\% of ASL \cite{Padden:2003} 
and is mainly used for lexical items that do not have their own ASL signs. Fingerspelled words are typically names, technical words, or words borrowed from another language, which makes its lexicon huge. Recognizing fingerspelling has great practical importance because fingerspelled words are often some of the most important context words.  

One problem in fingerspelling recognition is that relatively little curated labeled data exists, and even less data labeled at the frame level.  Recent work has obtained encouraging results using models based on neural network classifiers trained with frame-level labels \cite{scrf3}.  One goal of our work is to eliminate the need for frame labels.  In addition, most prior work has used hand-engineered image features, which are not optimized for the task.  A second goal is to develop end-to-end models that learn the image representation.
Finally, while labeled fingerspelling data is scarce, unlabeled fingerspelling or other hand gesture data is more plentiful.  Our final goal is to study whether such unlabeled data can be used to improve recognition performance.

We propose a model that jointly learns image frame features and sequence prediction with no frame-level labels. The model is composed of a feature learner and an attention-based neural encoder-decoder. The feature learner is based on an auto-encoder, enabling us to use unlabeled data of hand images (from both sign language video and other types of gesture video) in addition to transcribed data. We compare our approach experimentally to prior work and study the effect of model differences and of training with external unlabeled data. Compared to the best prior results on this task, we obtain 11.6\% and 4.4\% improvement respectively in signer-adapted and signer-independent letter error rates.\kl{changed 5.6\% to 4.4\% -- right?}

\section{Related Work}
\label{sec:relwork}

Automatic sign language recognition can be approached similarly to speech recognition, with signs being treated analogously to words or phones.  Most previous work has used approaches based on hidden Markov models (HMMs) \cite{hmm1, hmm2, hmm3, hmm4, hmm5}.  This work has been supported by the collection of several sign language video corpora, such as RWTH-PHOENIX-Weather \cite{rwth1, rwth2}, containing 
6,861 German Sign Language sentences, and the American Sign Language Lexicon Video Dataset (ASLLVD \cite{asllvd1, asllvd2, asllvd3}), containing video recordings of almost 3000 isolated signs.  

Despite the importance of fingerspelling in spontaneous sign language, there has been relatively little work explicitly addressing fingerspelling recognition.  Most prior work on fingerspelling recognition is focused on restricted settings. One typical restriction is the size of the lexicon. When the lexicon is fixed to a small size (20-100 words),
excellent recognition accuracy has been achieved \cite{reslexicon1,reslexicon2,reslexicon3}, but this restriction is impractical. 
For ASL fingerspelling, the largest available {\it open-vocabulary} dataset to our knowledge is the TTIC/UChicago Fingerspelling Video Dataset (Chicago-FSVid), containing 2400 open-domain word instances produced by 4 signers \cite{scrf3}, which we use here.  Another important restriction is the signer identity. 
In the signer-dependent setting, letter error rates below 10\% can be achieved for unconstrained (lexicon-free) recognition on the Chicago-FSVid dataset \cite{scrf1,scrf2,scrf3}; but the error rate goes above 50\% in the signer-independent setting and is around 28\% after word-level (sequence-level) \kl{added "sequence-level" since at this point I think "word-level" is a little unclear} adaptation \cite{scrf3}.  Large accuracy gaps between signer-dependent and signer-independent recognition have also been observed for general sign language recognition beyond fingerspelling \cite{hmm5}.

The best-performing prior approaches for open-vocabulary fingerspelling recognition have been based on HMMs or segmental conditional random fields (SCRFs) using deep neural network (DNN) frame classifiers to define features~\cite{scrf3}.  This prior work has largely relied on frame-level labels for training data, but these are hard to obtain.  In addition, because of the scarcity of data, prior work has largely relied on human-engineered image features, such as histograms of oriented gradients (HOG)~\cite{hog}, as the initial image representation.

Our goal here is to move away from some of the restrictions imposed in prior work.  To our knowledge, this paper represents the first use of end-to-end neural models for fingerspelling recognition without any hand-crafted features or frame labels, as well as the first use of external unlabeled video data to address the lack of labeled data.

\section{Methods}
\label{sec:methods}

\kl{added a few more words of context} Fingerspelling recognition from raw image frames, like many sequence prediction problems, can be treated conceptually as the following task: $(\mathbf{x}_1, \mathbf{x}_2,..., \mathbf{x}_S)$ $\rightarrow$   $(\mathbf{z}_1,\mathbf{z}_2,..., \mathbf{z}_S)$ $\rightarrow$ $(y_1, y_2,..., y_T)$, where $\{\mathbf{x}_i\}$, $\{\mathbf{z}_i\}$ ($1\leq i\leq S$) are raw image frames and image features, respectively, and $\{y_j\}$ ($1\leq j\leq T$) are predicted letters. Our model is composed of two main parts (which can be trained separately or jointly): \kl{added "which can..."} a feature extractor trained as an auto-encoder (AE) and an attention-based encoder-decoder for sequence prediction (see Figure \ref{struct}).  The attention-based model maps from $(\mathbf{z}_1,\mathbf{z}_2,..., \mathbf{z}_S)$ to $(y_1, y_2,..., y_T)$ and is similar to recent sequence-to-sequence models for speech recognition \cite{speech} and machine translation \cite{translation}.  For the feature extractor, we consider three types of auto-encoders: 

\begin{figure}[t]
\centering
\includegraphics[width=1.0\linewidth]{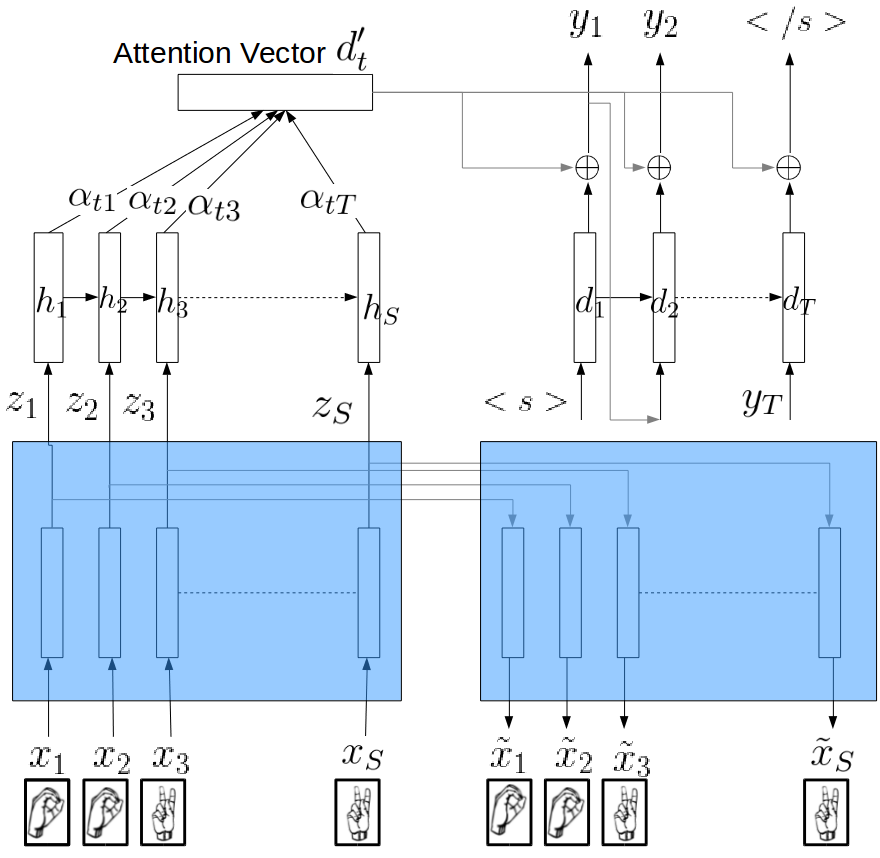}
\caption{\label{struct} Structure of the proposed model (blue region: auto-encoder, $\oplus$: concatenation).  The decoder component of the auto-encoder (the blue box on the right) is used only at training time. \kl{added sentence. also, is it possible to make the $\oplus$s in the figure match the one in the caption?}}
\end{figure}

\textbf{Vanilla Auto-Encoder (AE)} \cite{autoencoder}: A feedforward neural network consisting  of an encoder that maps the input (image) $\mathbf{x}\in \mathbb{R}^{d_x}$ to a latent variable $\mathbf{z}\in\mathbb{R}^{d_z}$, where $d_z < d_x$ and a decoder that maps $\mathbf{z}\in\mathbb{R}^{d_z}$ to output $\mathbf{\tilde{x}}\in \mathbb{R}^{d_x}$.  The objective is to minimize the reconstruction error $\mathcal{L}(\mathbf{x})=||\mathbf{x-\tilde{x}}||^2$ while keeping $d_z$ small. In our models we use multi-layer perceptrons (MLP) for both encoder and decoder.

\textbf{Denoising Auto-Encoder (DAE)} \cite{dae}: An extension of the vanilla auto-encoder where the input $\mathbf{x}$ at training time is a corrupted version of the original input $\mathbf{x}^\prime$. The training loss of the DAE is $\mathcal{L}(\mathbf{x};\mathbf{x}^\prime)=||\mathbf{x^\prime-\tilde{x}}||^2$

\textbf{Variational Auto-Encoder (VAE)} \cite{vae1,vae2} Unlike the vanilla and denoising auto-encoders, a variational auto-encoder models the joint distribution of the input $\mathbf{x}$ and latent variable $\mathbf{z}$: $p_{\theta}(\mathbf{x, z})= p_{\theta}(\mathbf{x|z})p_{\theta}(z)$. 
VAEs are trained by optimizing a variational lower bound on the likelihood $p(\mathbf{x})$:
\begin{equation}\label{eq_lowerbound}
\begin{split}
& \mathcal{L}(\mathbf{x})=-D_{KL}[q_{\phi}(\mathbf{z|x})||p_{\theta}(\mathbf{z})] + E_{q_{\phi}(\mathbf{z|x})}[\log\,p_{\theta}(\mathbf{x|z})] \\
\end{split}
\end{equation}
The two terms 
are the KL divergence between $q_{\phi}(\mathbf{z|x})$ and $p_{\theta}(\mathbf{z})$ and a reconstruction term $E_{q_{\phi}(\mathbf{z|x})}[\log\,p_{\theta}(\mathbf{x|z})]$. The prior $p_{\theta}(\mathbf{z})$ is typically assumed to be a centered isotropic multivariate Gaussian distribution $\mathcal{N}(\mathbf{0, I})$, and the posterior $q_{\phi}(\mathbf{z}|\mathbf{x})$ and conditional distribution $p_\theta(\mathbf{x}|\mathbf{z})$ are assumed to be multivariate Gaussians with diagonal covariance $\mathcal{N}(\pmb{\mu}_z, \pmb{\sigma}_z^2 \mathbf{I})$ and $\mathcal{N}(\pmb{\mu}_x, \pmb{\sigma}_x^2\mathbf{I})$. Under these assumptions, the KL divergence can be computed as

\begin{equation}\label{eq_kl}
D_{KL}[q_{\phi}(\mathbf{z|x})||p_{\theta}(\mathbf{z})]=\frac{1}{2}\displaystyle\sum_{d=1}^D(1+\log(\sigma_d^2)-\mu_d^2-\sigma_d^2)
\end{equation}
where $\pmb{\mu}_z=(\mu_1, ..., \mu_d)$ and $\pmb{\sigma}_z=(\sigma_1,..., \sigma_d)$ are approximated as the outputs of an MLP taking $\mathbf{x}$ as input.

Similarly to the AE and DAE, we use an MLP 
to model $\pmb{\mu}_x$ and $\pmb{\sigma}_x$. The loss of the VAE can thus be rewritten as
\begin{equation}\label{vae_loss}
\begin{split}
\mathcal{L}(\mathbf{x}) = & -\frac{1}{2}\displaystyle\sum_{d=1}^D(1+\log(\sigma_d^2)-\mu_d^2-\sigma_d^2) \\
& +\frac{1}{L}\displaystyle\sum_{l=1}^L{\log\,\mathcal{N}(\mathbf{x};\pmb{\mu}^{l}_x, \pmb{\sigma}^{l}_x)} \\
\end{split}
\end{equation}
where $L$ is a number of samples used to approximate the expectation in~\ref{eq_lowerbound}
(in practice we set $L=1$ as in prior work \cite{vae1}). $\pmb{\mu}_z$ is the feature vector $\mathbf{z}$ and $\pmb{\mu}_x$ serves the role of the reconstructed input $\tilde{\mathbf{x}}$ in Figure \ref{struct}.

{\bf RNN encoder-decoder:}
The latent variable sequence output by the auto-encoding module is fed into a long short-term memory (LSTM \cite{lstm}) recurrent neural network (RNN) for encoding:
$(\mathbf{z}_1, \mathbf{z}_2,..., \mathbf{z}_S)$ $\rightarrow$ $(\mathbf{h}_1, \mathbf{h}_2,..., \mathbf{h}_S)$. The LSTM states are fed into an RNN decoder that outputs the final letter sequence $(y_1, y_2,..., y_T)$. Attention \cite{attn} weights are applied to $(\mathbf{h}_1, \mathbf{h}_2,..., \mathbf{h}_S)$ during decoding in order to focus on certain chunks of image frames. If the hidden state of the decoder LSTM at time step $t$ is $\mathbf{d}_t$, the probability of outputting letter $y_t$, $p(y_t|y_{1:t-1}, \mathbf{z}_{1:T})$, is given by
\begin{equation}\label{eq_att}
\begin{split}
& \alpha_{it} = \text{softmax}(\mathbf{v}^{T}\tanh(\mathbf{W}_h \mathbf{h}_{i} + \mathbf{W}_d \mathbf{d}_t)) \\
& \mathbf{d}_t^\prime = \displaystyle\sum_{i=1}^S\alpha_{it}\mathbf{h}_{i} \\
& p(y_t|y_{1:t-1}, \mathbf{z}_{1:T}) = \text{softmax}(\mathbf{W_o}[\mathbf{d}_t;\mathbf{d^\prime}_t]+\mathbf{b}_o) \\
\end{split}
\end{equation}
and $\mathbf{d}_t$ is given by the standard LSTM update equation \cite{lstm}. 

The loss for the complete model is a multitask loss:
\begin{equation}\label{eq_totalloss}
\begin{split}
\mathcal{L}(\mathbf{x}_{1:T}, \mathbf{y}_{1:S})= & -\frac{1}{S}\displaystyle\sum_{j=1}^{S}\log\,p(y_j|y_{1:j-1}, \mathbf{z}_{1:T})\\
& +\frac{\lambda_{ae}}{T}\displaystyle\sum_{i=1}^T \mathcal{L}_{ae}(\mathbf{x}_{i})\\
\end{split}
\end{equation}
where $\mathcal{L}_{ae}(\cdot)$ is one of the losses of the AE, DAE or VAE, and $\lambda_{ae}$ measures the relative weight of the feature extraction loss vs. the prediction loss. 
 
\section{Experiments}
\label{experiments}

{\bf Data and experimental setup:} We use the TTIC/UChicago ASL Fingerspelling Video Dataset (Chicago-FSVid), which includes 4 native signers each fingerspelling 600 word instances consisting of 2 repetitions of a 300-word list containing common English words, foreign words, and names.\footnote{The recognition models do not use knowledge of the word list.} We follow the same preprocessing steps as in \cite{scrf3} consisting of hand detection and segmentation, producing 347,962 frames of hand regions. In addition, we also collect extra unlabeled handshape data consisting of 65,774 ASL fingerspelling frames from the data of \cite{extra1} and 63,175 hand gesture frames from \cite{extra2}. We chose these external data sets because they provide hand bounding boxes; obtaining additional data from video data sets without bounding boxes is possible (and is the subject of future work), but would require hand tracking or detection. Despite the smaller amount of external data, and although it is noisier than the UChicago-FSVid dataset (it includes diverse backgrounds), it provides examples of many additional individuals' hands,
which is helpful for signer-independent recognition.  All image frames are 
scaled to $64\times 64$ before feeding into the network.

\begin{figure*}[t]
\centering
\includegraphics[width=1.0\textwidth]{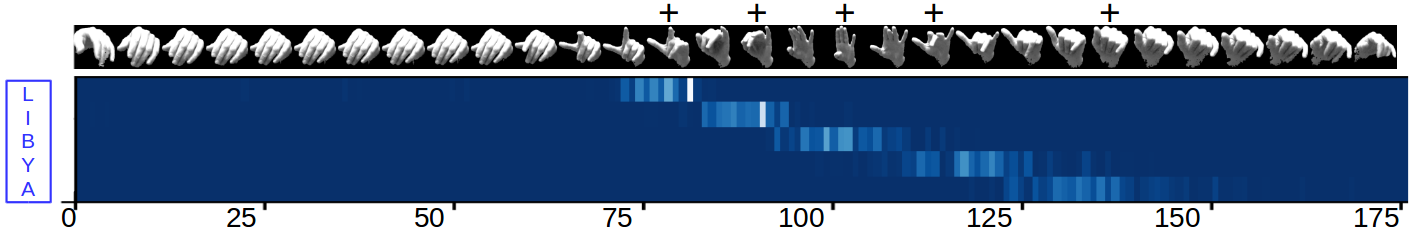}
\caption{\label{att_fig}Attention visualization for the example word ``LIBYA".  Colors correspond to the attention weights $\alpha_{it}$ in Equation \ref{eq_att}, where $i$ and $t$ are column and row index, respectively.
Lighter color corresponds to higher value. At the top are subsampled image frames for this word; 
frames with a plus (+) are the ones with highest attention weights, which are also the most canonical handshapes in this example.  (Alignments between image frames and attention weights are imperfect due to frame subsampling effects.)} 
\end{figure*}

Our experiments are done in three settings: signer-dependent (SD), signer-independent (SI) and signer-adapted (SA). We use the same setup as in \cite{scrf1,scrf2,scrf3}, reviewed here for completeness. For the SD case, models are trained and tested on a single signer's data.  The data for each signer is divided into 10 subsets for k-fold experiments.  
80\%, 10\% and 10\% of the data are respectively used as train, validation, and test sets in each fold.  8 out of 10 possible folds are used (reserving 20\% of the data for adaptation), and the reported result is the average letter error rate (LER) over the test sets in those 8 folds. For the SI case, we train on three signers' data and test on the fourth. 
For the SA case, the model is warm-started from a signer-independent model and fine-tuned with 20\% of the target signer data.\footnote{In previous work on signer adaptation \cite{scrf2,scrf3}, multiple approaches were compared and this was the most successful one.}  10\% of the target signer data is used for hyperparameter tuning and test results are reported on the rest. 
Previous work has considered two types of adaptation, using frame-level labels (alignments) for adaptation data or only word-level labels; here we only consider word-level adaptation. 

{\bf Model details} The auto-encoder consists of a 2-layer MLP encoder and 2-layer MLP decoder with 800 ReLUs in each hidden layer and dimensionality of the latent variable $\mathbf{z}$ fixed at 100. Weights 
are initialized with Xavier initialization \cite{xavier}. Dropout is added between layers at a rate of 0.8.\footnote{Dropout rate refers to the probability of retaining a unit.}
For the sequence encoder and decoder, we use a one-layer LSTM RNN with hidden dimensionality of 128 and letter embedding dimensionality of 128. 
We use the Adam optimizer \cite{adam} with initial learning rate 0.001, which is decayed by a factor of 0.9 when the held-out accuracy stops increasing. Beam search is used for decoding; 
the effect of beam width will be discussed later. The default value for $\lambda_{ae}$ in the multitask loss function (Equation \ref{eq_totalloss}) is 1, but it can be tuned. The model is trained first with the unlabeled data, using only the auto-encoder loss, and then the labeled data using the multitask loss.  We also experimented with iteratively feeding unlabeled and labeled data, but this produced worse performance.

\vspace{-.05in}
\subsection{Baselines}

\begin{table}[tb]
\centering
\small
\begin{tabular}{|llrll|}\hline
  & Model & \shortstack{SD} & \shortstack{SI} & \shortstack{SA} \\ \hline
1 & Best prior results~\cite{scrf3} & $\textbf{7.6}^a$ & $55.3^b$ & $27.9^c$ \\
\hline
 2 & HOG + enc-dec & 11.1 & 50.3 & 29.1 \\
 3 & CNN + enc-dec & 9.1 & 50.7 & 28.7 \\
 4 & DNN + enc-dec & 9.9 & 50.9 & 29.3 \\ \hline
 5 & CNN + enc-dec+ & 10.7 & 50.4 & 29.2  \\
 \hline
 6 & E2E CNN-enc-dec & 11.8 & 48.4 & 27.5 \\ 
 7 & E2E DNN-enc-dec & 12.1 & 47.9 & 26.9 \\ \hline 
 8 & AE + enc-dec & 21.7 & 61.2 & 40.3 \\
 9 & DAE + enc-dec & 17.5 & 56.4 & 34.4 \\
 10 & VAE + enc-dec & 18.8 & 58.2 & 37.8 \\ \hline
 11 & E2E AE-enc-dec & 11.8 & 48.1 & 30.2 \\
 12 & E2E DAE-enc-dec & 11.9 & 45.0 & 28.9 \\
 13 & E2E VAE-enc-dec & 10.6 & 43.8 & 23.8 \\ \hline
 14 & E2E AE-enc-dec* & 10.0 & 47.3 & 29.2 \\
 15 &E2E DAE-enc-dec* & 9.5 & 44.3 & 27.2 \\
 16 & E2E VAE-enc-dec* & 8.1 & \textbf{43.7} & \textbf{23.5} \\ \hline
\end{tabular}
\caption{\label{tab_res} Letter error rates (\%) of different models.  SD: signer-dependent, SI: signer-independent, SA: signer-adapted.  Model names with an asterisk (*) and a plus (+) use extra unlabeled hand image data and augmented data respectively.  Best prior results are obtained with SCRFs ($^a$ = 2-pass SCRF, $^b$ = rescoring SCRF, $^c$ = first-pass SCRF).}
\end{table}

We compare the performance of our approach with the best prior published results on this dataset, obtained with various types of SCRFs and detailed in~\cite{scrf3}.  These prior approaches are trained with frame-level labels.
In addition to the results in \cite{scrf3}, we consider the following extra baselines. 

{\bf Baseline 1 (HOG + enc-dec)}: We use a classic hand-engineered image descriptor, histogram of oriented gradient (HOG~\cite{hog}), and directly feed it into the attention encoder-decoder. We use the same HOG feature vector as in \cite{scrf3}. This baseline allows us to compare engineered features with features learned by a neural network.

{\bf Baseline 2 (CNN + enc-dec, DNN + enc-dec)}:
A CNN or DNN frame classifier is trained using frame letter labels, and its output (pre-softmax layer) is used as the feature input $\mathbf{z}$ in the attention encoder-decoder.  The classifier network is not updated during encoder-decoder training.  
This baseline tests whether frame-level label information is beneficial for 
the neural encoder-decoder. The input for both CNN and DNN are the $64\times 64$ image pixels concatenated over a 21-frame window (10 before and 10 following the current frame). The DNNs have three hidden layers of sizes 2000, 2000 and 512. Dropout is added between layers at a rate of 0.6. The CNNs are composed of (in order) 2 convolutional layers, 1 max-pooling layer, 2 convolutional layers, one max-pooling layer, 3 fully connected layers, and 1 softmax layer. The stride in all convolutional layers is 1 and the filter sizes are respectively: $3\times 3\times 21\times 32$, $3\times 3\times 32\times 32$, $3\times 3\times 32\times 64$, $3\times 3\times 64\times 64$. Max-pooling is done over a window of size $2\times 2$ with stride 2. Finally the fully connected layers are of sizes 2000, 2000 and 512. Dropout at a rate of 0.75 and 0.5 is used 
for the convolutional and fully connected layers, respectively. The fully connected layers in both CNN and DNN have rectified linear unit (ReLU) \cite{nair2010rectified} \kl{replaced relu citation; please double-check} activation functions. Training is done via stochastic gradient descent with initial learning rate 0.01, which is decayed by a factor of 0.8 when the validation accuracy decreases after the first several epochs.  The network structural parameters (number and type of layers, number of units, \emph{etc.}) are tuned according to the validation error, and the above architectures are the best ones in our tuning.

{\bf Baseline 3 (E2E CNN/DNN+enc-dec)}:  End-to-end version of {\emph CNN/DNN + enc-dec}. In this baseline, the CNN/DNN parameters are learned jointly with the encoder-decoder and no frame labels are used. 

{\bf Baseline 4 (AE/DAE/VAE + enc-dec)}: Separate training of auto-encoder and encoder-decoder modules, each with its own loss.  Baselines 3 and 4 are used to study the effectiveness of end-to-end training.

\begin{figure*}[h]
\includegraphics[width=1.0\textwidth]{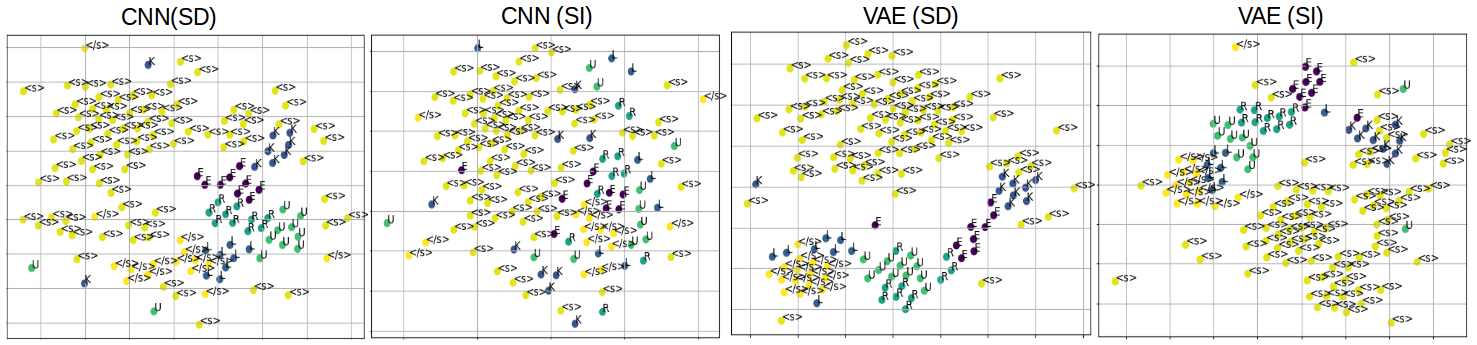}
\caption{\label{tsne} Visualization via a 2-D t-SNE embedding \cite{tsne} of image frame features extracted in the end-to-end VAE model and the CNN classifier for example word ``KERUL" in the signer-dependent (SD) and signer-independent (SI) settings.}
\end{figure*}

\subsection{Results}

The overall results are shown in Table~\ref{tab_res}.  Our main findings are as follows:

{\bf Best-performing model:} The proposed end-to-end model, when using a VAE and the external unlabeled data (line 16), achieves the best results in the signer-independent (SI) and signer-adapted (SA) cases, improving over the previous best published results by 11.6\% and 4.4\% absolute, respectively.  In all of our end-to-end models (11-16), the VAE outperforms the AE and DAE.  In the signer-dependent case, our best model is 0.5\% behind the best published SCRF result, presumably because our model is more data-hungry and the SD condition has the least training data.  

{\bf Encoder-decoders vs.~prior approaches:} More generally, models based on RNN encoder-decoders (lines 2-16) often outperform prior approaches (line 1) in the SI and SA settings but do somewhat worse in the signer-dependent case.  
We visualize the attention weights in Figure \ref{att_fig}. The frame corresponding to the canonical handshape often has the highest attention weight. The alignment between the decoder output and image frames is generally monotonic, though we do not use any location-based priors.

{\bf The effect of end-to-end training:} We measure the effect of end-to-end training vs.~using frame labels by comparing the separately trained CNN/DNN + enc-dec (lines 3-4) with their end-to-end counterparts (lines 6-7), as well as separately trained AEs (lines 8-10) vs.~their E2E counterparts (lines 11-13). We find that separate training of a frame classifier can improve error rate by about $2\%$ in the signer-dependent setting, but in the other two settings, end-to-end models trained without frame labels consistently outperform their separate training counterparts. Features learned by a frame classifier seem to not generalize well across signers.  The non-end-to-end AE-based models do much worse than their E2E counterparts, presumably because the feature extractor does not get any supervisory signal.  We visually compare the features of each image frame trained through an end-to-end model vs.~a frame classifier via t-SNE~\cite{tsne} embeddings (Figure \ref{tsne}). We find that both feature types show good separation in the SD setting, but in the SI setting the end-to-end VAE encoder-decoder has much clearer clusters corresponding to letters.

\begin{figure}[tb]
\includegraphics[width=1.0\linewidth]{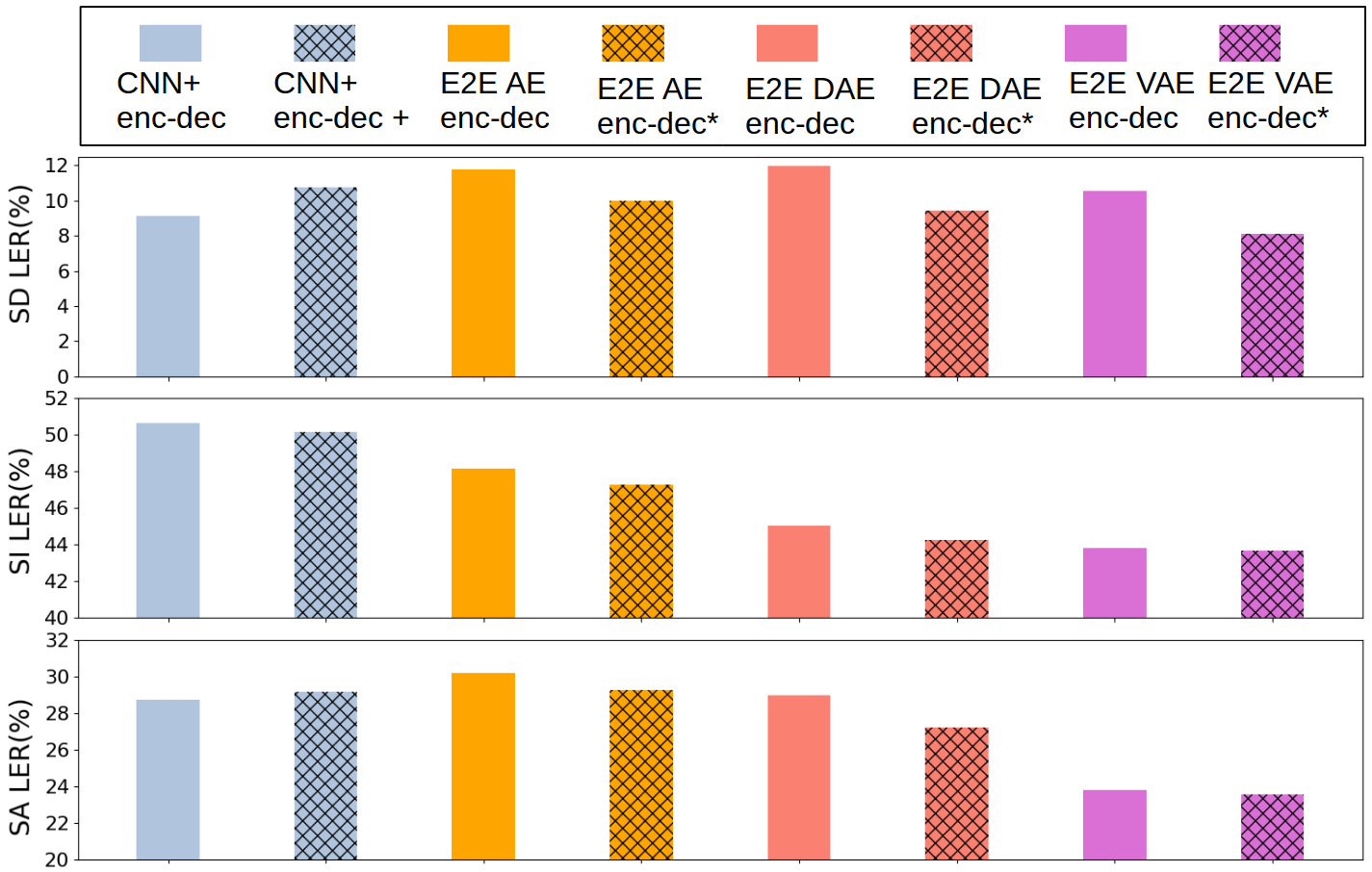}
\caption{\label{fig_res_extra}Comparison of different models with and without extra data (*: with external data, +: with augmented data).}
\end{figure}

{\bf Does external unlabeled data help?} 
The extra data gives a consistent improvement for all three auto-encoding models in all settings (lines 11-13 vs.~14-16 and Figure~\ref{fig_res_extra}). 
The average accuracy improvements for the three settings are respectively 2.1\%, 0.6\%, and 0.7\%. 
The SI and SA improvement is smallest for the best (VAE-based) model, but the overall consistent trend suggests that we may be able to further improve results with even more external data.  The improvement is largest in the SD setting, perhaps due to the relatively larger amount of extra data compared to the labeled training data.

\begin{figure*}[htp]
\centering
\includegraphics[width=1.0\textwidth]{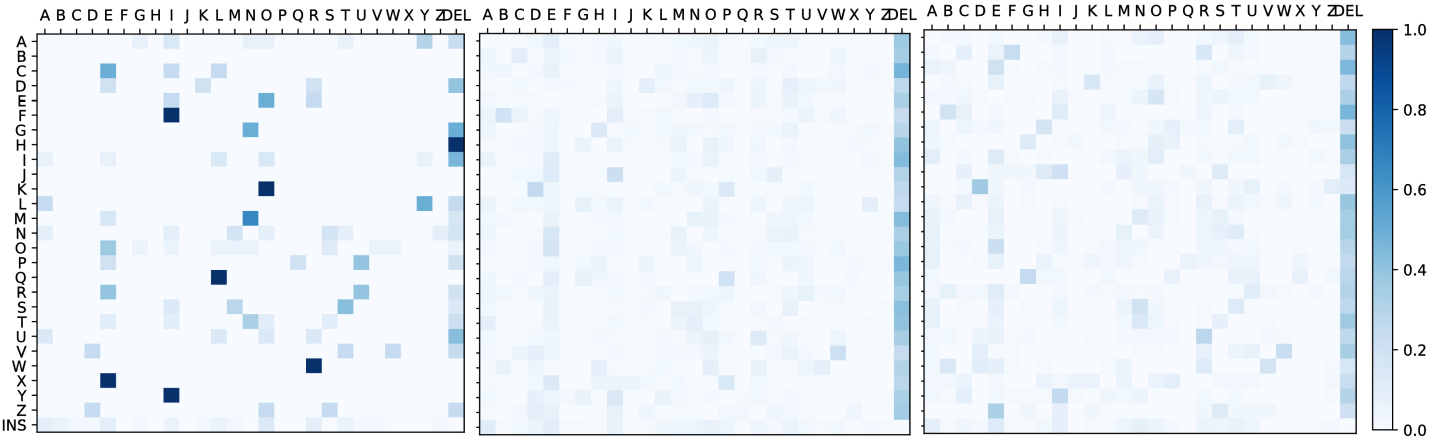}
\caption{\label{confmat}Letter confusion matrix under three settings (from left to right: signer-dependent (SD), signer-independent (SI) and signer-adapted (SA)).  The color in each cell corresponds to the empirical probability of predicting a hypothesized letter (horizontal axis) given a certain ground-truth letter (vertical axis).  The diagonal in each matrix has been removed for visual clarity. \kl{added 2 sentences}}
\end{figure*}

{\bf Would data augmentation have the same effect as external data?}
We compare the extra-data scheme to classic data augmentation techniques \cite{data_aug}, which involve adding replicates of the original training data with geometric transformations applied.  We perform the following transformations: scaling by a ratio of 0.8 and translation in a random direction by 10 pixels, rotation of the original image at a random angle up to 30 degrees both clockwise and counterclockwise. We generate augmented data with roughly the same size as the external data (960 word and 168,950 frames) and then train the CNN/DNN + enc-dec model (with frame labels).  The results (Figure \ref{fig_res_extra} and Table~\ref{tab_res} line 5 vs.~3) show that data augmentation hurts performance in the SD and SA settings and achieves a 0.3\% 
improvement in the SI setting. 
We hypothesize that the extra unlabeled hand data provides a richer set of examples than do the geometric transformations of the augmented data.

\begin{figure}[htb]
\centering
\includegraphics[width=1.0\linewidth]{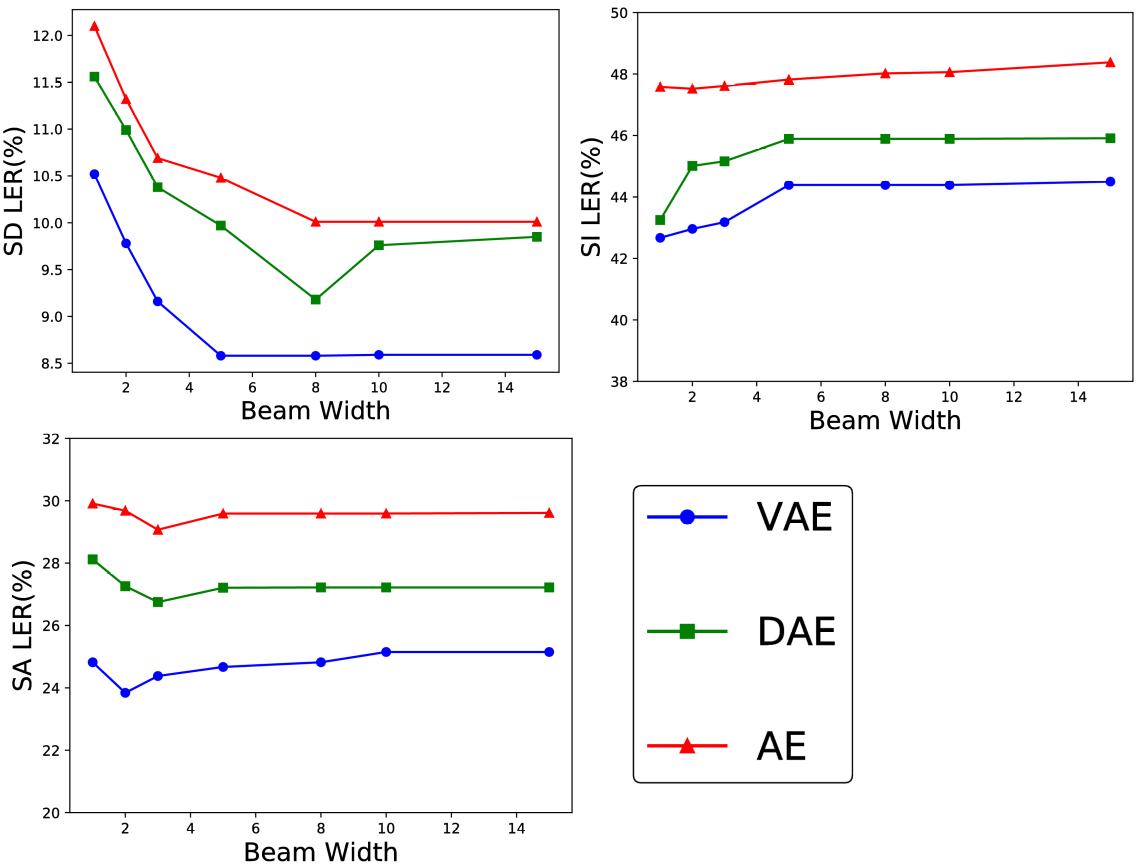}
\caption{\label{beam_curve}Letter error rate (\%) with different beam widths in signer-dependent (SD), signer-independent (SI) and signer-adapted (SA) settings.}
\end{figure}

{\bf Effect of beam width:}
We analyze the influence of beam width on error rates, shown in Figure \ref{beam_curve}. Beam search is important in the SD setting. In this setting, the main errors are substitutions among similar letter handshapes (like e and o), as seen from the confusion matrix in Figure \ref{confmat}. Using a wider beam can help catch such near-miss errors.  However, in the SI and SA settings, there are much more extreme differences between the predicted and ground-truth words, evidenced by the large number of deletion errors in Figure \ref{confmat}. Therefore it is hard to increase accuracy through beam search.  
Some examples of predicted words are listed in Table \ref{tab_beam}.

\begin{table}[h]
\small
\begin{tabular}{|c|c@{\hskip 0.1in}c@{\hskip 0.1in}c@{\hskip 0.1in}c|}\hline
& B=1 & B=3 & B=5 & \small{Ground-truth} \\ \hline
\multirow{2}{*}{\shortstack{\small{SD} \\ \;}} & \small{FIRSWIUO} & \small{FIREWIUE} & \small{FIREWIRE} & \small{FIREWIRE} \\
 &  \small{NOTEBEEK} & \small{NOTEBOOK} & \small{NOTEBOOK} & \small{NOTEBOOK} \\ \hline
\multirow{2}{*}{\shortstack{\small{SI} \\ \;}} & \small{AAQANNIS} & \small{AOQAMIT} & \small{AOQUNIR} & \small{TANZANIA} \\
  & \small{POPLDCE} & \small{POPULCE} & \small{POPULOE} & \small{SPRUCE} \\ \hline
\end{tabular}
\caption{\label{tab_beam}Example outputs with different beam sizes in signer-dependent and signer-independent settings.}
\end{table}

\section{Conclusion}

We have introduced an end-to-end model for ASL fingerspelling recognition that jointly learns an auto-encoder based feature extractor and an RNN encoder-decoder for sequence prediction. The auto-encoder module enables us to use unlabeled data to augment feature learning.  We find that these end-to-end models consistently improve accuracy in signer-independent and signer-adapted settings, and the use of external unlabeled data further slightly improves the results. Although our model does not improve over the best previous (SCRF-based) approach in the signer-dependent case, this prior work required frame labels for training while our approach does not.  Future work includes collecting data ``in the wild" (online) and harvesting even more unlabeled data.

\section*{Acknowledgements}

We are grateful to Greg Shakhnarovich and Hao Tang for helpful suggestions and discussions. This research was funded by NSF grant 1433485.



\bibliographystyle{IEEEbib}
\bibliography{strings,refs}

\end{document}